# Application of Random Forest and Support Vector Machine for Investigation of Pressure Filtration Performance, a Zinc Plant Filter Cake Modeling

Masoume Kazemi[1], Davood Moradkhani[1✉], Alireza A. Alipour[2]

## Abstract

The hydrometallurgical method of zinc production involves leaching zinc from ore and then separating the solid residue from the liquid solution by pressure filtration. This separation process is very important since the solid residue contains some moisture that can reduce the amount of zinc recovered. This study modeled the pressure filtration process through Random Forest (RF) and Support Vector Machine (SVM). The models take continuous variables (extracted features) from the lab samples as inputs. Thus, regression models namely Random Forest Regression (RFR) and Support Vector Regression (SVR) were chosen. A total dataset was obtained during the pressure filtration process in two conditions:  1) Polypropylene (S1) and 2) Polyester fabrics (S2). To predict the cake moisture, solids concentration (0.2 and 0.38), temperature (35 and 65 °C), pH (2, 3.5, and 5), pressure, cake thickness (14, 20, 26, and 34 mm)， air-blow time (2, 10 and 15 min) and filtration time were applied as input variables. The models' predictive accuracy was evaluated by the coefficient of determination (called $R^2$) parameter. The results revealed that the RFR model is superior to the SVR model for cake moisture prediction.

1 Zanjan University, Department of Materials Science and Engineering
2 Institute for Advanced Studies in Basic Sciences, Department of Computer Science
✉ Corresponding author: Faculty of Engineering Zanjan University, Zanjan, Iran
E-mail address: moradkhani@znu.ac.ir✉ Corresponding author: Faculty of Engineering Zanjan University, Zanjan, Iran

## 1. Introduction

Metal production and processing require balancing economic benefits, resource efficiency, and environmental impacts [1]. More than 85% of zinc nowadays is produced by hydrometallurgical processes [2]. Figure 1 shows the flow sheet of the zinc hydrometallurgical process used in Iran, which differs from other methods in the purification steps of nickel, cobalt, and cadmium due to environmental limitations [3].

The zinc plant feed contains impurities such as iron, nickel, cobalt, and cadmium, which are removed in different stages of the process (Figure 1). The residues also contain moisture with cations and anions, leading to metal losses. Zinc Leaching Plant Residue (ZPR) is the most abundant residue and by-product of zinc production. Its amount varies with the feed grade and is approximately 3 to 6 times the zinc output [4].

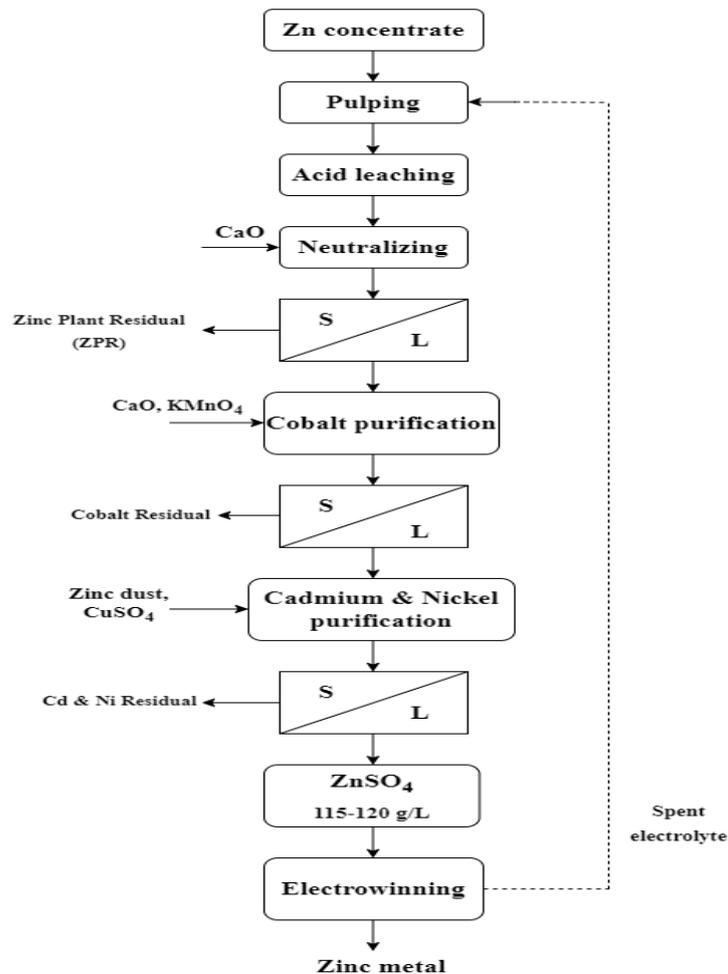

Fig. 1 Zinc plant flowsheet [3]

Zinc production from Zinc Plant Residue (ZPR) has been investigated by various researchers. E. Vahidi et al. [5] reported that Di-2-EthylHexyl Phosphoric Acid (D2EHPA) was an effective zinc extraction and separation solvent. Yunpeng Du et al. [6] evaluated the environmental impact of zinc recovery from ZPR and the toxicity of the residues using the Toxicity Characteristic Leaching Procedure (TCLP). E. Guler et al. [7] studied the influence of sulfating roasting conditions on zinc's metal solubility and extraction efficiency from ZPR. Several studies have also reported high zinc recovery rates from ZPR. M. Deniz et al. [8] achieved 86% zinc recovery by using a hydrometallurgical process. Huan et al. [9] combined reduction roasting, acid leaching, and magnetic separation to obtain 61.38% zinc recovery. The main challenges for zinc recovery from ZPR are the undissolved zinc, the hydroxide precipitation, and the moisture content in the residue. These factors depend on the process and equipment parameters used in zinc production. Zinc losses in ZPR can occur due to non-dissolution, precipitation, or moisture. Non-dissolution is affected by the comminution, degree of freedom of zinc, and leaching conditions of the zinc minerals. Precipitation and moisture are related to the neutral leaching and filtration stages of the process. The filtrate of ZPR contains 100-150 g/L of zinc, which means that the residue has a high zinc content in its moisture. To enhance zinc recovery, it is essential to optimize the reduction of the moisture content in the residue. However, previous studies have not tested the optimal filtration conditions of ZPR on a pilot scale.

Researchers have used various mathematical tools for data processing and computer modeling. However, Artificial Intelligence (AI) based prediction models have recently become an important tool for forecasting in many areas. This process involves three steps: data collection and data preparation, creating Machine Learning (ML) algorithms to support advanced analytics, and using these algorithms to predict outcomes. A clear objective of this process is to build a model that can analyze the output accurately and reliably. Therefore, adequate and reliable data are essential for this purpose.

Modeling is the process of selecting relevant algorithms, training them from training data, and obtaining accurate predictions. Machine learning can be expansively divided into three categories: supervised learning, unsupervised learning, and reinforcement learning [10-14]. In supervised learning, the training data have a definite output and a corresponding label. In unsupervised learning, the training data have no specific output or no label. In reinforcement learning, the algorithm learns from its experience to map the situation to action, and in consequence, maximizes a numerical reward signal. This means that the algorithm receives feedback from its actions and adjusts its behavior accordingly.

Machine learning is fully transforming many industries into automation. Also, it has many applications in the field of material science, especially in modeling and optimization. For example, Jiayang Dai et al. (2020) [15] implemented a spatiotemporal model based on KL-MS-LLSSVM to control the ferrous ion concentration in the goethite process during iron removal. Moreover, the results demonstrated that the model could reduce the consumption of oxygen and zinc oxide. This is important because the electro-winning processes consume a lot of energy and need to be more efficient. Since the consumption of energy during electro-winning processes is substantial and must be reduced, Xiongtao Shi et al. (2020) [16] proposed a Deep Deterministic Policy Gradient (DDPG) learning controller. Similarly, XIONG Jie et al. (2020) [17] provided another approach that used random forest regression to predict the mechanical properties of structural materials. A different approach is the Back Propagation (BP) neural network model that utilized iron powder smelting temperature, calcium oxide, sodium carbonate, and coke as input variables to limit the discharges of pollutants from the secondary lead smelting industry [18]. Another reliable method is the feed-forward back-propagation ANN that obtained the effect of solid concentration, initial pH, time, and inoculum percent on the dissolution of Cu, Mo, and Re from mineral molybdenite via meso-acidophilic bioleaching [19]. In the current work, the genetic algorithm caused an optimum combination of the operational parameters for attaining the maximum recoveries of Cu, Mo, and Re by the neural network [20]. Another application of machine learning in material science is the protection and prediction of the thickening and filtration processes. For example, Hui Li et al. (2016) [21] used a Bayesian network to protect against the abnormity that occurs in the thickening process of gold hydrometallurgy. As a result, the safe control was fully automated and the abnormity was eliminated. Avalos et al. (2020) [26] studied several predictive methods: polynomial regression, k-nearest neighbor, support vector machine, multilayer perceptron, long short-term memory, and gated recurrent units. They concluded that the workflow has the potential of being extended to any other temporal and multivariate mineral processing datasets. This research aims to use machine learning to reduce the moisture content of ZPR. For this purpose, we conducted pilot scale experiments to develop support vector machine and random forest models for pressure filtration of the zinc process. Then, we presented feasible solutions for predicting cake moisture and evaluated the relative importance of the selected variables, such as pressure, time, and temperature. Finally, we compared the forecasting performance of the two models and developed the conclusions of the work.

## 2. Methods

In this section, we will review the machine learning methods used in this research, namely random forest and support vector machine. First, we will explain what machine learning is and how it works. Machine learning is a time-saving way that learns automatically from data and predicts based on data [22,23]. It can be divided into three broad categories: supervised, unsupervised, and reinforcement learning [24]. Second, we will focus on the supervised learning algorithms that are the most widespread method of machine learning in material science. These algorithms use labeled data to train and test the models. Third, we will introduce random forest and support vector machine as two of the AI tools that are used to predict the moisture of leaching filter cake. These tools are appropriate for modeling and simulation in materials science for the design and development of processes.

### 2.1. Random Forest

In this section, we will explain the random forest algorithm. It is a supervised machine-learning algorithm that constructs decisions for classification and regression problems. Random Forest Regression (RFR), which is a variant of RF that is used for regression problems, where the goal is to predict a continuous value instead of a discrete class. It has been shown that random forest has better regression accuracy than other models. In other words, a random forest creates numerous decision trees in the training phase and takes the average output value of all individual trees as the outcome. Moreover, random forest is a reliable method for quantifying the significant variables of the class.

The main concept of the random forest model is to combine multiple classification regression trees with weaker performance as a forest using specific rules and make predicting results by voting among all decision trees in the forest [25]. The random forest model f is composed of decision trees and is defined as Eq. (1).

$$\{h(X, \theta_k), k = 1, 2, \cdots, n\} \quad (1)$$

where X and $\theta_k$ describe the input and random vector, in turn. Distribution of $\theta_k$ occurred independently within the k-th decision tree. The input vector X includes up to Y categories. Considering the input vector X and output vector Y, the edge function is calculated by Eq. (2).

$$K(X, Y) = a_k I[h(X, \theta_k) = Y] - \max_{j \neq Y} a_k I[h(X, \theta_k) = j] \quad (2)$$

where j is a type of training set and $a_k$ is the average function. A higher edge function corresponds closely to the classification correctness. The generalization error of the RF model is formulated as follows in Eq. (3).

$$E^* = P_{X,Y}(K(X,Y) < 0) \tag{3}$$

The RF model convinces the following theorems if there is a high number of decision trees in the forest

**Theorem 1.** By increasing the number of trees for each $\theta_k$, converging the error of the RF model ($E^*$) to zero is more probable which is expressed as Eq. (4).

$$P_{X,Y}\left(P_\theta(h(X,\theta) = Y) - \max_{j \neq Y} P_\theta(h(X,\theta) = j) < 0\right) \to 0 \tag{4}$$

Based on the first theorem, despite increasing the number of trees, overfitting does not happen as the generalization error of an RF model. Nevertheless, it does not tend to have a particular value.

**Theorem 2**. To calculate the upper bound on the RF model generalization error, the represented formula is used by Eq. (5)

$$E^* \leq \frac{\bar{\rho}(1-s^2)}{s^2} \tag{5}$$

where $\rho$ is the average correlation coefficient and s defines the average strength of the tree. The theorem explains that reducing the tree correlation and increasing the strength of a single tree is the reason for declining the upper bound on the generalization error to be effectively controlled. Practically, just the two parameters, the number three and the split characteristic number mtry have to be included in forecasting, since it is the number of them selected per tree that affects forecasting performance.

### 2.2. Support Vector Machine

Support Vector Machine (SVM) is another supervised learning method that we used in this research. It is a machine learning algorithm based on statistical learning theory [10] that performs regression and classification analysis. By introducing Vapnik's ε-insensitive loss function, SVM has been widened to resolve the regression problems, which is called Support Vector Regression (SVR). Statistical and mathematical learning theory has shown that the SVR method approximates an unknown function by mapping input data into a high-dimensional feature space via a nonlinear mapping function. Then, a linear problem is formed in this feature space.

The SVR aims for discovering a functional relationship $f(x)$ between input data $x_i$ and output data $z_i$ at most $\epsilon$ deviation as flat as possible by the presumption of the joint distribution $P$ of $(x, z)$ is completely unknown. The application of kernel-trick is to model nonlinear relationships, furthermore, to convert the complicated nonlinear problem into a simple linear problem a mapping $\Phi$ is introduced. The primary concepts of the standard SVR algorithm with ε-insensitive loss function are explained below.

A sample set $D = \{(x_i, z_i)\}$ is assumed where $x_i \in R$ is the input values and $z_i \in R$ are the related output values for $i = 1, 2, \ldots, N$ where $N$ is the samples number. Therefore, the whole problem can be formulated as the convex optimization problem represented in Eq. (6).

minimize
$$\frac{1}{2}\|\omega^2\| + C \sum_{i=1}^{1}(\xi_i + \xi_i^*)$$

(6)

subject to
$$\begin{cases} <\omega \cdot \Phi(x)> + b - z_i \leq \varepsilon + \xi_i \\ z_i - <\omega \cdot \Phi(x)> - b \leq \varepsilon + \xi_i^* \\ \xi_i, \xi_i^* \geq 0 \end{cases}$$

where $C$ is a constant which is known as the penalty factor to take control of the tradeoff between the smoothness of $f(x)$ and the tolerance to errors over the $\epsilon$, and the slack variables $\xi_i$, $\xi_i^*$ term is the indication of the amount of difference between the estimated value and the target value.

## 3. Experimental Method

Zinc concentrate that satisfies the needs of the leaching process for this study was obtained from Calcimin company, Zanjan, Iran, with 25% moisture. To prepare the concentrate for the leaching process, the zinc concentrate was crushed by a jaw crusher in the Material and Engineering Department at Zanjan University. Figure 2 shows the filtration stage that produced the Zinc Plant Residual (ZPR).

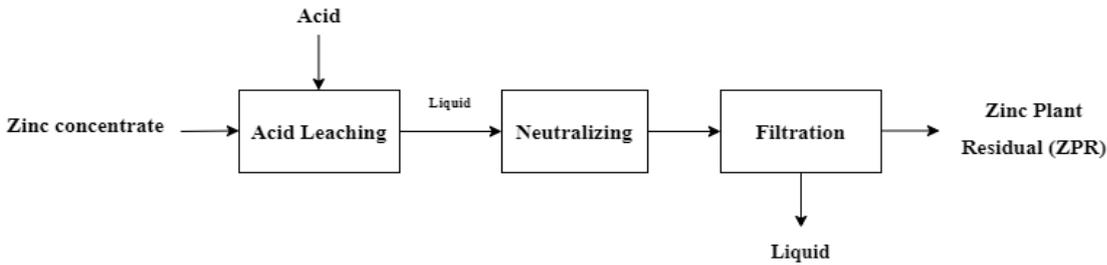

Fig. 2 Configuration of ZPR production experiments

The moisture of the filter cake with 25% moisture was determined by the filtration of the leaching process. A plate filter press performed 288 experiments (144 with polypropylene and 144 with polyester fabric) at different levels of solid concentration (SC), temperature (Tem), pH, air-blow time (ABT), cake thickness (CT), and filtration time (FT).

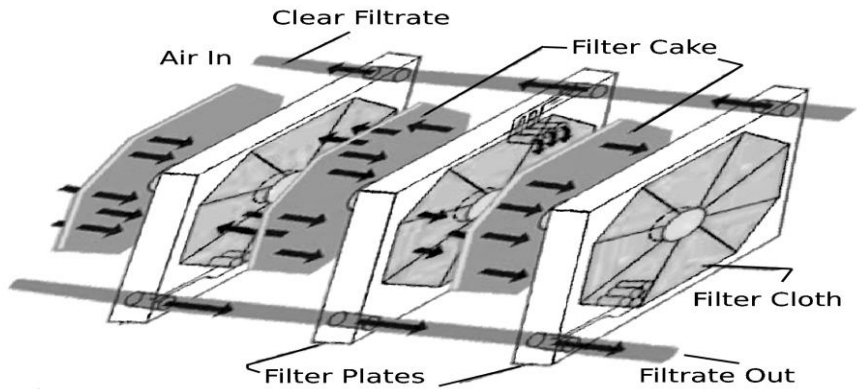

Fig. 3 Schematic of the cake thickness

First, the ball mill crushed the zinc concentrate to a powder. Then, a 200L tank with a mechanical stirrer and a controller unit leached 37.5 Kg samples of zinc concentrate in 125 and 62.5 L water at a solid: liquid ratio of 0.2 and 0.38 g/L, respectively. The leach solution was heated at two temperatures (35 °C and 65 °C) and the pH was regulated by $H_2SO_4$ (98%) and lime. Next, the solution was filtered from the solid material. Four sizes of plate filter presses (14, 20, 26, and 34 mm) were used to vary the cake thickness, as shown in Figure 3. The air was blown in for 2,10 and 15 minutes, respectively. Then, the known weight of ZPR was dried at 110-120 °C for 8h in the oven to measure the filter cake's moisture. Table 1 shows the ranges of these parameters.

Table 1. Input parameter ranges for the filtration process

| Parameter | Level | | | |
|---|---|---|---|---|
| | 1 | 2 | 3 | 4 |
| Solid concentration (g/L) | 0.2 | 0.38 | - | - |
| Temperature (°C) | 35 | 65 | - | - |
| pH | 2 | 3.5 | 5 | - |
| Air-blow time (min) | 2 | 10 | 15 | - |
| Cake thickness (mm) | 14 | 20 | 26 | 34 |

## 4. Development, Training, and Testing Models

The goal of this section is to test the prediction accuracy of the proposed models using various experiments. Two datasets are used to evaluate the performance of the proposed forecasting model; they are obtained from pressure filtration experiments with different types of fabric, namely polypropylene, and polyester. The first dataset (S1) has 144 data points from polypropylene fabric, and the second dataset (S2) has 144 data points from polyester fabric. The datasets consist of seven input parameters: solids concentration, temperature, pH, pressure, filtration time, air-blow time, and cake thickness.

Table 2. Summary of descriptive statistics for the experimental data (S1)

|  | Solids Concentration, % | Temperature, °C | pH | Pressure, kPa | Air-blow Time min | Cake Thickness, mm | Filtration Time min | Cake Moisture, % |
|---|---|---|---|---|---|---|---|---|
| Minimum | 0.20 | 32.00 | 2.09 | 150 | 2.00 | 14.00 | 7.34 | 26.09 |
| 1st Quartile | 0.20 | 35 | 2.11 | 150 | 2.00 | 15.50 | 9.25 | 31.94 |
| Median | 0.29 | 49.50 | 3.52 | 150 | 10.00 | 23.50 | 12.00 | 33.11 |
| Mean | 0.29 | 50.00 | 3.57 | 150 | 9.00 | 23.00 | 11.82 | 33.17 |
| 3rd Quartile | 0.38 | 63.75 | 4.91 | 150 | 15.00 | 32.00 | 14 | 34.47 |
| Maximum | 0.38 | 68.00 | 5.67 | 150 | 15.00 | 34.00 | 16.00 | 39.76 |

The output parameter is considered the moisture of the filter cake. The Support Vector Regression (SVR) and Random Forest Regression (RFR) models are used to predict the cake moisture of the filter cake. The datasets are split into training and validation sets randomly, with 70% (100 data points) for training and 30% (44 data points) for validation. Feature selection is performed on the datasets to find the optimal RBF kernel for the SVR models based on experience. Normalization is applied to the training dataset to avoid numerical issues; the data are scaled to the range [0, 1]. Table 2 and Table 3 show the descriptive statistics for S1 and S2, respectively.

Table 3. Summary of descriptive statistics for the experimental data (S2)

|  | Solids Concentration, % | Temperature, °C | pH | Pressure, kPa | Air-blow Time min | Cake Thickness, mm | Filtration Time min | Cake Moisture, % |
|---|---|---|---|---|---|---|---|---|
| Minimum | 0.20 | 32.00 | 2.00 | 150 | 2.00 | 14.00 | 6.50 | 24.45 |
| 1st Quartile | 0.20 | 33.25 | 2.10 | 150 | 2.00 | 15.50 | 10.00 | 31.73 |
| Median | 0.29 | 50.00 | 3.45 | 150 | 10.00 | 23.00 | 10.00 | 33.47 |
| Mean | 0.29 | 49.42 | 3.47 | 150 | 9.00 | 23.50 | 9.96 | 33.57 |
| 3rd Quartile | 0.38 | 64.75 | 4.74 | 150 | 15.00 | 32.00 | 10.38 | 35.24 |
| Maximum | 0.38 | 67.00 | 5.10 | 150 | 15.00 | 34.00 | 11.50 | 40.94 |

To examine the prediction performance of the SVR and RFR model, their forecasting errors in terms of Coefficient of determination ($R^2$), Mean Squared Error (MSE), and Mean Absolute Error (MAE) are obtained which is defined as given by Equation (6), (7), and (8), respectively:

$$R^2 = 1 - \frac{\sum_{i=1}^{n}(y-\hat{y})^2}{\sum_{i=1}^{n}(y)^2} \tag{6}$$

$$MSE = \frac{1}{n} \sum_{i=1}^{n}(y-\hat{y})^2 \tag{7}$$

$$MAE = \frac{1}{n} \sum_{i=1}^{n}|y-\hat{y}| \tag{8}$$

where y represents an actual value, $\hat{y}$ is the predicted value, and n is the number of selected forecasting points.

## 5. Results and Discussion

The cake moisture of zinc slurry is an important factor in the leaching process of zinc production, as it affects the amount of zinc that is dissolved or undissolved in water. Therefore, reducing the cake moisture in the first step of filtration is crucial. In this study, Support Vector Regression (SVR) and Random Forest Regression (RFR) have been developed to predict the cake moisture of zinc slurry.

### 5.1. Comparison of Machine Learning Models

The performance of the support vector regression and random forest regression models was assessed on 288 datasets of cake moisture from the pressure filtration process using two types of fabric: polypropylene (S1) and polyester (S2). The datasets were randomly split into training and validation sets, with 100 and 44 data points each.

Figure 4 illustrates the predicted cake moisture versus the actual cake moisture by the RFR model for both S1 and S2 data, respectively. As can be seen in Figure 4, the experimental data are on the x-axis, while the predicted values obtained from the RFR model are on the y-axis. Each point on the scatter plot corresponds to a sample, and its coordinates are the actual and predicted values of the output variable. For example, sample #44 has an actual value is 0.41 and the predicted value is 0.38, the point will be at (0.41, 0.38) on the plot. Therefore, it shows a good fit between the actual and predicted values using the RFR model.

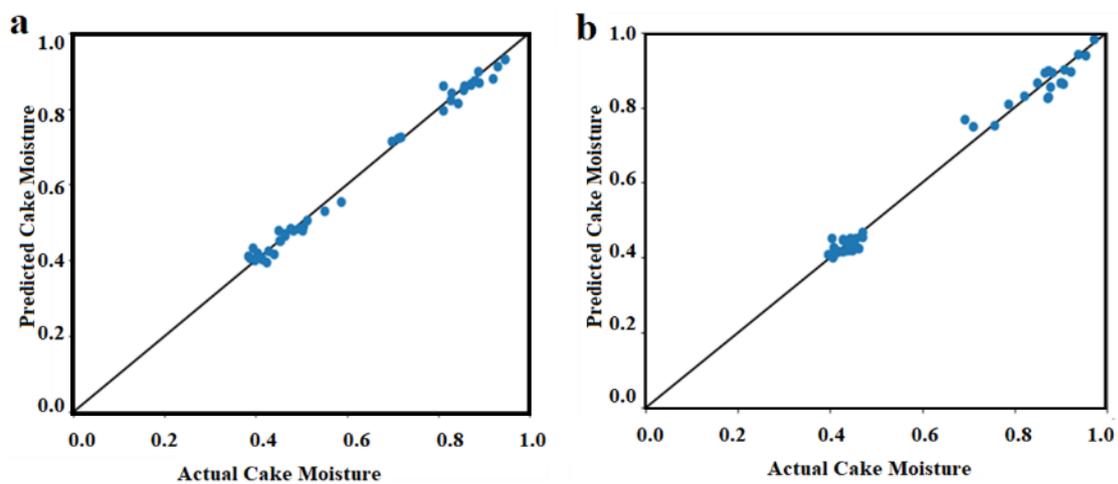

Fig. 4 Comparison of actual cake moisture with the predicted cake moisture for RFR model: (a) S1; (b) S2

The graphs in Figure 5 compare the performance of the SVR model in predicting the target values from the experimental data. In Figure 5, the x-axis and y-axis represent the same variables as the RFR model, but using the SVR model. There is a considerable deviation of the predicted values from the actual values using the SVR model in Figure 5. To reduce the error rates of both models, the data points should be closer to the line of equality (y=x).

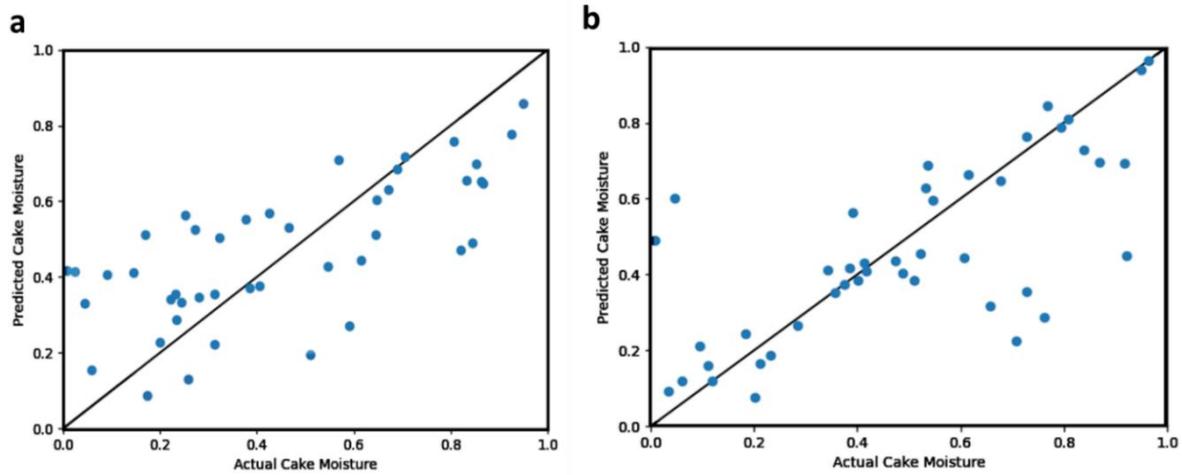

Fig. 5 Comparison of actual cake moisture with the predicted cake moisture for the SVR model: (a) S1; (b) S2

Table 4 shows the statistical evaluation parameters of the RFR model for predicting the values of S1 and S2. The RFR model had a high prediction accuracy, as indicated by the coefficient of determination ($R^2$) of 0.991 and 0.987 for S1 and S2, respectively. The RFR model also had low forecasting errors, as indicated by the Mean Squared Error (MSE) of $4.398 \times 10^{-08}$ and $8.636 \times 10^{-08}$ and the Mean Absolute Error (MAE) of 0.00015 and 0.00022 for S1 and S2, respectively.

Table 4: The statistical evaluation parameters of the RFR-based model

|  | S1 | S2 |
|---|---|---|
| $R^2$ | 0.991 | 0.987 |
| MSE | $4.398 \times 10^{-08}$ | $8.636 \times 10^{-08}$ |
| MAE | 0.00015 | 0.00022 |

However, the SVR model performed poorly on both S1 and S2 datasets, as shown by its low $R^2$ value of 0.48 and its high MSE (0.16 and 0.12) and MAE (0.039) values. On the other hand, the RFR model achieved a much higher $R^2$ value than the SVR model, as shown in Table 5.

Table 5: The statistical evaluation parameters of the SVR-based model.

|  | S1 | S2 |
|---|---|---|
| $R^2$ | 0.48 | 0.48 |
| MSE | 0.16 | 0.12 |
| MAE | 0.039 | 0.039 |

The SVR and RFR models were applied to predict the cake moisture of zinc slurry based on data from pressure filtration. The line y=x represents a perfect fit, where the model predicts the same value as the actual one for every sample. The model performance is better when the points are closer to this line and worse when the points are farther from this line. The RFR model showed a high degree of fit between the actual and forecasted values, as illustrated in Figure 4. The forecasted values closely followed the actual values, and its forecasting errors in terms of the MSE and MAE values were very low, indicating the high accuracy of the RFR model. In contrast, the SVR model produced forecasts that deviated significantly from the actual values, as shown in Figure 5. The SVR model also had the highest MSE and MAE values among the models. Moreover, Figure 4 shows that the RFR model had a better alignment with the y=x line than the SVR model, which means that the RFR model had fewer prediction errors and better generalization ability. These results confirmed the superior performance of the random forest model over the support vector model in predicting cake moisture, based on the statistical evaluation parameters. Therefore, the RFR model was the most suitable model for this task.

### 5.2. Relative Importance of Input Variables

This research used support vector regression and random forest regression to evaluate the influence of different parameters on the pressure filtration of zinc slurry. The input variables were ranked by their importance to the prediction model. Selecting the most relevant variables can reduce the problem of high dimensionality.

The relative importance of the variables obtained by the RFR model is shown in Figure 6. Furthermore, pH and temperature are the least important variables for both S1 and S2 according to the RFR model, while filtration time has opposite effects for S1 and S2, respectively.

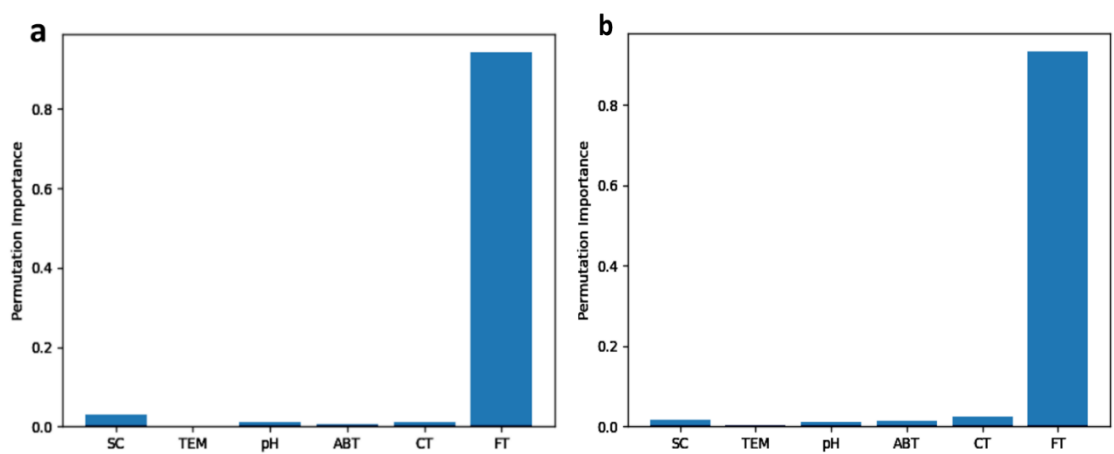

Fig. 6 Relative importance of input variables from RFR model: (a) S1, (b) S2

Figure 7 shows how important each variable is for the SVR model. In Figure 7 (a), pH and temperature have positive effects, while solid concentration has a negative effect on S1. In addition, temperature is the most influential variable for S2, and filtration time is a minor factor as shown in Figure 7 (b).

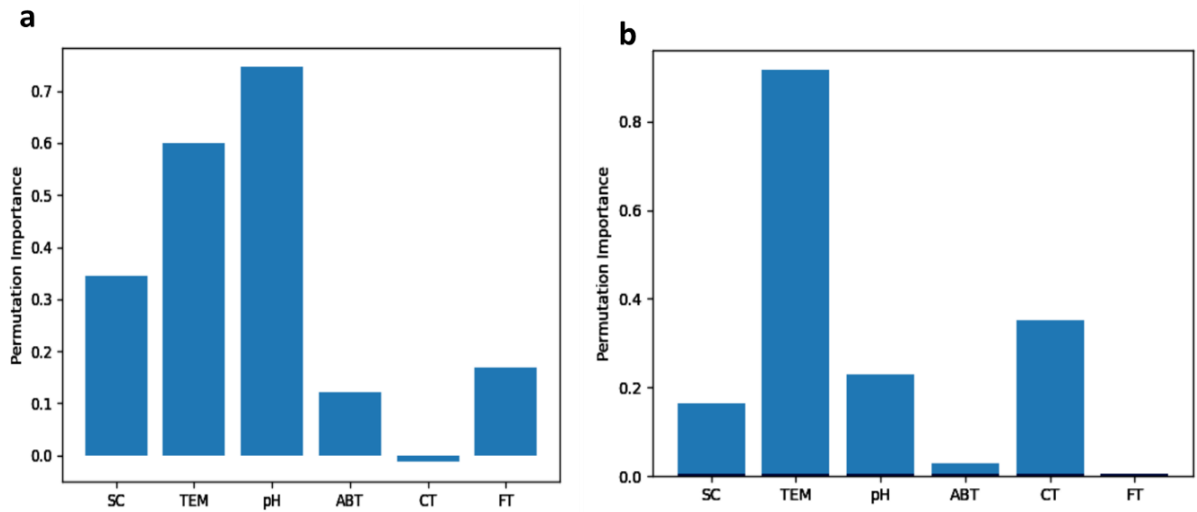

Fig. 7 Relative importance of input variables from the SVR model: (a) S1, (b) S2

## 6. Conclusion

In this study, we developed two machine learning models, Support Vector Regression (SVR) and Random Forest Regression (RFR), to forecast the cake moisture of zinc slurry. We used two different scenarios, S1 and S2, to train and test our models on real-world data. We evaluated the performance of our models using three metrics: Mean Squared Error (MSE), Mean Absolute Error (MAE), and Coefficient of determination ($R^2$). The results showed that RFR outperformed SVR in all metrics and scenarios. The MSEs for prediction of cake moisture by RFR and SVR models in the testing (validation) stage were $6.636 \times 10^{-08}$ and $8.636 \times 10^{-08}$, 0.16, and 0.12 for S1 and S2, respectively. Also, MAEs for cake moisture prediction were obtained as 0.00015 and 0.00022, 0.039 and 0.039 for S1 and S2, respectively by above mentioned models. Furthermore, the $R^2$ values between the experimentally measured and calculated values of cake moisture using RFR were higher than those using SVR, indicating a better fit and agreement. In conclusion, according to the evaluation parameters, the comparison results of RFR with the SVR model verify that the random forest regression model is an excellent model than the support vector regression model in terms of prediction performance, accuracy, and generalization ability. Consequently, RFR can be a very powerful tool in pressure filtration of hydrometallurgy processes.

## Acknowledgments

The authors would like to express their sincere gratitude to the research and technology deputy of the University of Zanjan for providing them with the opportunity and facilities to use the hydrometallurgical pilot plant. Also, we would like to thank Dr. Ebrahim Ansari, Assistant Professor of the Computer Science and Information Technology Department for their invaluable help in the process of developing this paper.